\documentclass{article}
\usepackage[margin=1in]{geometry}
\usepackage{amsmath,amsfonts}
\usepackage{algorithmic}
\usepackage{algorithm}
\usepackage{array}
\usepackage[caption=false,font=normalsize,labelfont=sf,textfont=sf]{subfig}

\usepackage{graphics}
\usepackage{multirow}
\usepackage{bm}
\usepackage{caption}
\usepackage[switch, columnwise]{lineno}

\usepackage{textcomp}
\usepackage{stfloats}
\usepackage{float}
\usepackage{url}
\usepackage{booktabs}
\usepackage{verbatim}
\usepackage{graphicx}
\usepackage{biblatex}
\usepackage{xcolor}
\usepackage{tikz}
\usepackage{xspace}

\usetikzlibrary{bayesnet}
\usetikzlibrary{arrows}
\usetikzlibrary{positioning}
\bibliography{references}

\begin{document}
\def\imagesize{40}
\def\metersperpixel{120 m}
\def\resolutionmeters{4.8 km}
\def\areameters{23.04 km$^2$}
\def\pccnn{\texttt{PixelConstrainedCNN}\xspace}
\def\sccar{\texttt{SCCAR}\xspace}

\title{Deep autoregressive modeling for land use and land cover} 

\author{Christopher Krapu, Department of Civil and Environmental Engineering, Duke University\\ 
Mark Borsuk, Department of Civil and Environmental Engineering, Duke University \\
Ryan Calder, Department of Population Health Sciences, Virginia Tech}

\maketitle

\begin{abstract}
    Land use / land cover (LULC) modeling is a challenging task due to long-range dependencies between geographic features and distinct spatial patterns related to topography, ecology, and human development. We identify a strong connection between modeling spatial patterns of land use and the task of image inpainting from computer vision, and conduct a study of a modified PixelCNN architecture with approximately 19 million parameters for modeling LULC. Compared with a benchmark spatial statistical model, we find that the former is capable of capturing much richer spatial correlation patterns such as roads and water bodies but does not produce a calibrated predictive distribution, suggesting the need for further tuning. We find evidence of predictive underdispersion with regard to important ecologically-relevant land use statistics such as patch count and adjacency, which can be mitigated to some extent by manipulating sampling variability.
\end{abstract}

\section{Introduction}

The combination of widely available satellite imagery and land-use classification models has led to the development of national- and global-scale land use / land cover (LULC) databases \cite{vogelmann2001, gong2013}. These data are recorded as categories such as \emph{forest}, \emph{pasture}, \emph{open water} and are frequently derived from satellite or aerial imagery. Additionally, they are referenced to a grid with pixel dimensions ranging from 10 m \cite{brown2022} to 30 m \cite{wickham2023} to 1000 m \cite{latham2014}. 
Some of these data products are available for multiple points in time \cite{brown2022}, though we do not consider the temporal dimension in this work. 
Spatial patterns of land use / land cover data arise from a rich interplay between climate, topography, soil, and human intervention among many other factors. Consequently, modeling the spatial correlation patterns between different classes of LULC can be highly challenging; the characteristic size and shape of a patch of forest is likely to be quite different from that of a pond or town due to differing processes. Despite this challenge, the investigation into LULC models is a actively pursued topic due to their utility in ecology \cite{nunes2022}, hydrology \cite{carlson2000} and urban planning \cite{landis1995}. 

In general, the task addressed by LULC models is to predict the class or proportion of different classes represented at a spatial location either in isolation or jointly with other locations. Inputs for this modeling process can include co-located data such as elevation or soil maps as well as top-down constraints such as population. Existing data on the spatial and statistical distribution of nearby land cover is also commonly used as an input. Existing LULC models, as reviewed by \cite{ren2019}, fall into a few broad classes such as cellular automata (CA)\cite{white2000}, statistical models \cite{gao2019}, single-pixel neural network models \cite{pijanowski2002}, and agent-based models (ABM) \cite{parker2001}. The modeling approach adopted in this work is distinct in that it is calibrated to a large number of training examples, unlike most ABM and CA methods, allows for much more flexible outputs than a standard statistical approach, and is capable of generating extensive patches of land use not just individual pixels.

The intent of this work is to enhance the existing set of LULC modeling strategies by including deep generative models through an autoregressive \pccnn architecture \cite{vandenoord2016b}, making use of pixel-level constraints \cite{dupont2019} to leverage information otherwise inaccessible to an unmodified PixelCNN model for prediction. Our particular use case is motivated by the need for flexible predictive conditional distributions for LULC over relatively large regions ($N\approx 10^3$ pixels). We aim to move beyond two-point or simpler multipoint statistical models and adopt models with millions of parameters for this task. Specifically, we consider a section of pixels and aim to produce a distribution of possible scenarious for spatial regions with areas of several square kilometers or more using data at much higher spatial resolution. In simpler terms, we are to create a range of likely alternative patterns of LULC for an area given data on the nearby surroundings. This challenge is similar to that of assessing ecosystem services as in \cite{lawler2014}; we wish to speculate on the possible LULC patterns that might be present if a certain landscape feature like a state park or military base had not been established.\\

To contextualize our research within the broader machine learning community, we note that this problem presents several unique challenges. (1), we need models appropriate for categorical imagery as opposed to the more common grayscale or multichannel RGB images usually modeled as a range of continuous or ordinal values. (2), as each image represents a small area of a larger expanse, there is the possibility to use an enormous amount of additional spatial context, such as summaries of adjacent areas, soil type maps, topographical maps, and nearby points of interest in modeling. (3), we require \emph{probabilistic} estimates of the missing pixels and place a great value on predictive distributions that are properly calibrated according to relevant summary statistics. PixelCNN architecture easily solves item \#1 while item \#2 is a topic for future study; most of our effort focused on item \#3, verifying if deep autoregressive models satisfied our needs. Therefore, our research revolves around these questions:

\begin{itemize}
    \item Are deep autoregressive models capable of capturing the complex patterns of LULC with long-range spatial dependencies?
    \item Are the resulting joint predictive distributions over sets of multiple pixels properly calibrated for applications that follow?
\end{itemize}

In the rest of this section, we clarify the problem statement and provide more detail relevant to LULC modeling and deep generative models. We utilize a coarsened version of the USA National Land Cover Database files for the continental USA from 2019 for our analyses; this and other methodological specifics are elaborated in the Methods section, with our discoveries presented in the Results and Discussion sections.

\section{Data and Methods}
As the primary goal of this work is to assess generative machine learning models for producing LULC images, we considered several model variants including initial experiments with variational autoencoders (VAE) as well as generative adversarial networks (GAN); we found sampling image completions with the former to be more computationally intensive than the approach adopted here. 
We also found VAE-derived image completions to have less detail, though we did not use recently developed VAE architectures as presented in \cite{razavi2019}. 
GANs are challenging to use with discrete data as their training requires backpropagation through the generated data, for which the standard gradient cannot be produced in this case, though some workarounds for this issue exist \cite{kusner2016}. Additionally, we found autoregressive models to produce acceptably detailed LULC images, in part due to their decomposition of the image generation task into many per-pixel iterations as opposed to methods requiring a single pass through the network. 

Within the context of generative modeling, an autoregressive model treats an image as a sequence of pixel intensities, modeling each pixel intensity as a conditional distribution given the preceding pixel intensities. 
Given an image $\bm{x}$ represented as a two-dimensional array of size $N=H \times W$ with a single discrete channel where pixel values are in the set $\left\{0,..., K-1\right\}$, the joint probability of the image can be factorized as a product of conditional probabilities. Assuming a raster scan order in which we proceed left-to-right followed by top-to-bottom through the $N$ pixels of the image, the autoregressive model factorization can be written as: $p(\bm{x}) = \prod_{i=1}^N f(\bm{x}_{i} | \bm{x}_{<i})$. In this expression, $p(\bm{x}_{i} | \bm{x}_{<i})$ represents the conditional probability of pixel at location $i$ given all preceding pixels, denoted by $\bm{x}_{<i}$, in the sequence. This probability is often modeled by a parametric function, for instance a deep neural network $f_\theta$ with parameters denoted by $\theta$. These parameters are learned from training data, typically using a maximum likelihood objective function. Examples of models within this class include PixelRNN \cite{vandenoord2016b}, PixelCNN \cite{oord2016}, PixelCNN++ \cite{salimans2017}, PixelSNAIL \cite{chen2018}, and ImageGPT\cite{chen2020}.  We note that these network designs typically do not allow for conditioning on out-of-order pixels; this is accomplished using a constrained PixelCNN network described in a following section. 

Despite the advantages of autoregressive models for high-quality image generation and inherent ability to perform probabilistic sampling, relatively few studies exist at the intersection of autoregressive models and geospatial data. These include the use of PixelCNN variants for anomaly detection in overhead imagery \cite{montserrat2020} as well as the generation of street networks with a transformer-based autoregressive model \cite{birsak2022}. 
\section{Methods}

\subsection{Data}

\subsubsection*{National Land Cover Database}
The National Land Cover Database is a data product generated by the US Geological Survey in conjunction with the Multiresolution Land Characteristics Consortium. It is derived from satellite imagery obtained under the Landsat program and is available at the same spatial resolution as Landsat imagery, albeit at reduced temporal resolution. Per-pixel classifications are produced with a decision tree classifier \cite{jin2019} applied to Landsat imagery. A review of similar or related LULC data products is presented in \cite{garcia-alvarez2022}. Other work involving generative modeling and NLCD data include the incorporation of NLCD as a weak label for generation of higher resolution land cover maps via diffusion models \cite{rolf2022, graikos2023}. 

\begin{figure}[H]
    \centering
    \includegraphics[width=0.7\linewidth]{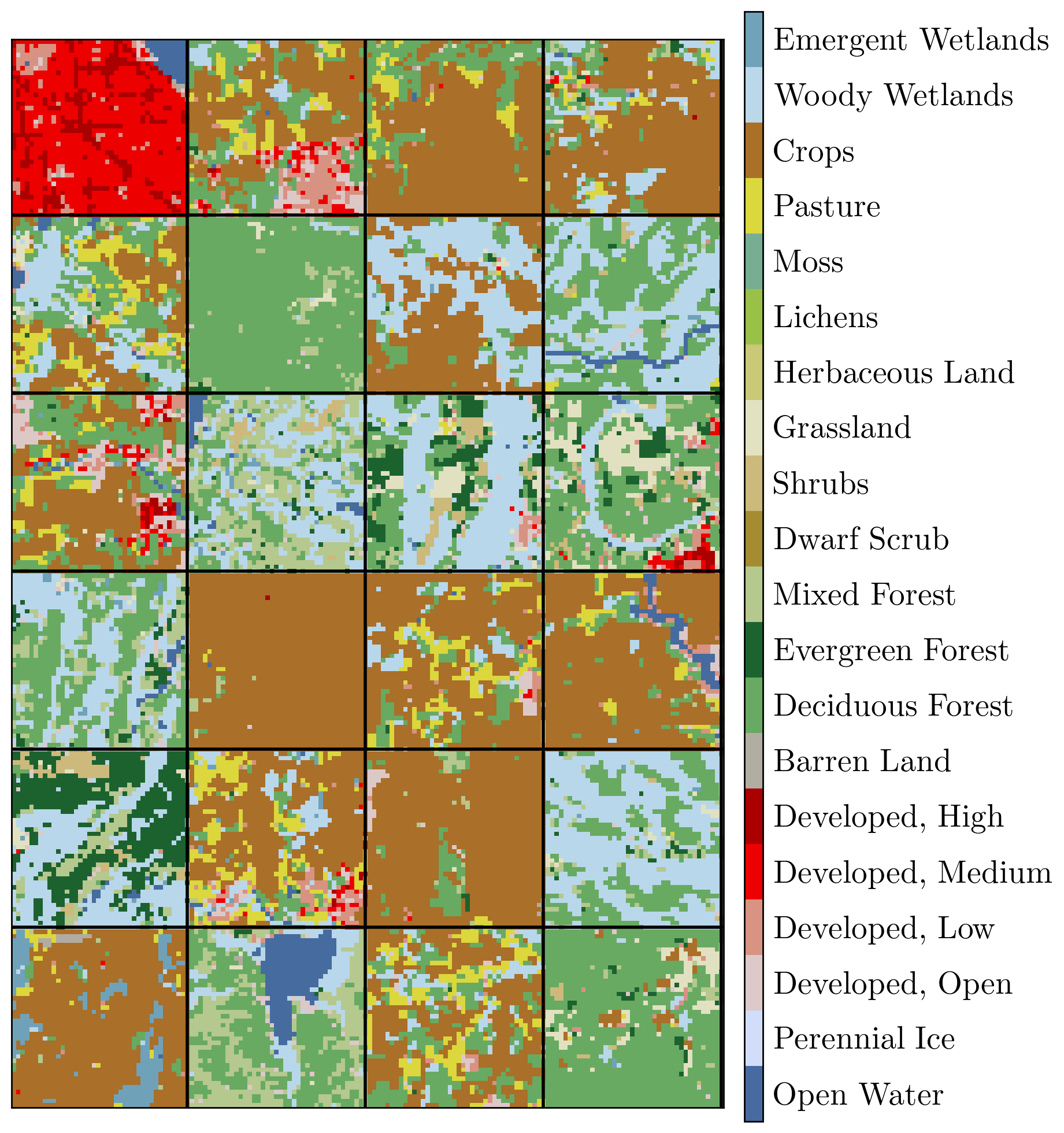}
    \caption{\textbf{Examples of training data at a $\imagesize\times \imagesize$ resolution}. Common cover co-occurrence patterns include pasture and cropland, woody wetlands and forest, as well as high/medium/low levels of urban development. Each image covers \areameters.}
    \label{fig:training}
\end{figure}

NLCD data have been regenerated at multiple times over the past two decades, including data for 2011, 2016, and 2019. We use the 2019 version in this work and aggregate it by a factor of 3 by representing each larger pixel as the dominant land cover type in the $3 \times 3$ patch at the original resolution. Thus, each image of size $\imagesize\times \imagesize$ is \resolutionmeters per side and has an area \areameters. We apply this coarsening to allow for easier modeling of occluded regions of the same spatial scale as a moderately-sized urban development or land parcel; this case study examines the particular case for the spatial extent of an existing military base. Figure \ref{fig:training} shows selections of this data in the format used for modeling. 

We constructed a training data set of 100,000 images in the geographic region contained within the rectangle bounded by latitude ranges from 42.331 N. - 43.64 N. and longitude ranges from 83.11 W. to 86.21 W (Figure \ref{fig:likelihood}). This region overlaps the states of Wisconsin, Illinois, Michigan, Indiana, and Ohio, and covers several predominant land cover types including cropland, forest, and urban developments. We focus on this spatial subset to reduce training computing costs and to use a data distribution with more narrow support but a rich collection of LULC spatial patterns including wetlands, riparian areas, roads, urban development, forest, and cropland. Further work with larger models could involve training against the entire NLCD dataset or similar globally available LULC data. While some of the NLCD categories were not present in the data considered in this work, we retain all 20 categories in our implementation for off-the-shelf compatibility with data sets comprising all categories. Each image covers an area of \areameters with height and width dimensions of \metersperpixel. We also constructed a test data set of 10,000 images in the same region. To avoid including a large number of noninformative data points, we did not sample any images with more than 50\% pixels in the \emph{open water} class. 
\subsection{Models}

\subsubsection*{Pixel-constrained CNN}
PixelCNN, an autoregressive model, captures the joint distribution of an image $\bm{x}$'s pixels by factoring it into a product of conditional distributions. Each pixel's value is conditioned on the values of all preceding pixels in a predefined order. In the context of our application, where each pixel takes one of $K = 20$ land use categories, we model the conditional probability of a pixel's category given all preceding pixels using a softmax function. Under this model, the log-probability of image $\bm{x}$ is $\log p(\bm{x}) = \sum_{i=1} f_{\theta}(\bm{x}_{i}|\bm{x}_{<i})$, with the summation order following the same order as the pixel conditioning order. However, the PixelCNN architecture as presented in \cite{vandenoord2016} does not condition on pixels outside this ordering.

\begin{figure}[H]
    \centering
    \includegraphics[width=1.0\linewidth]{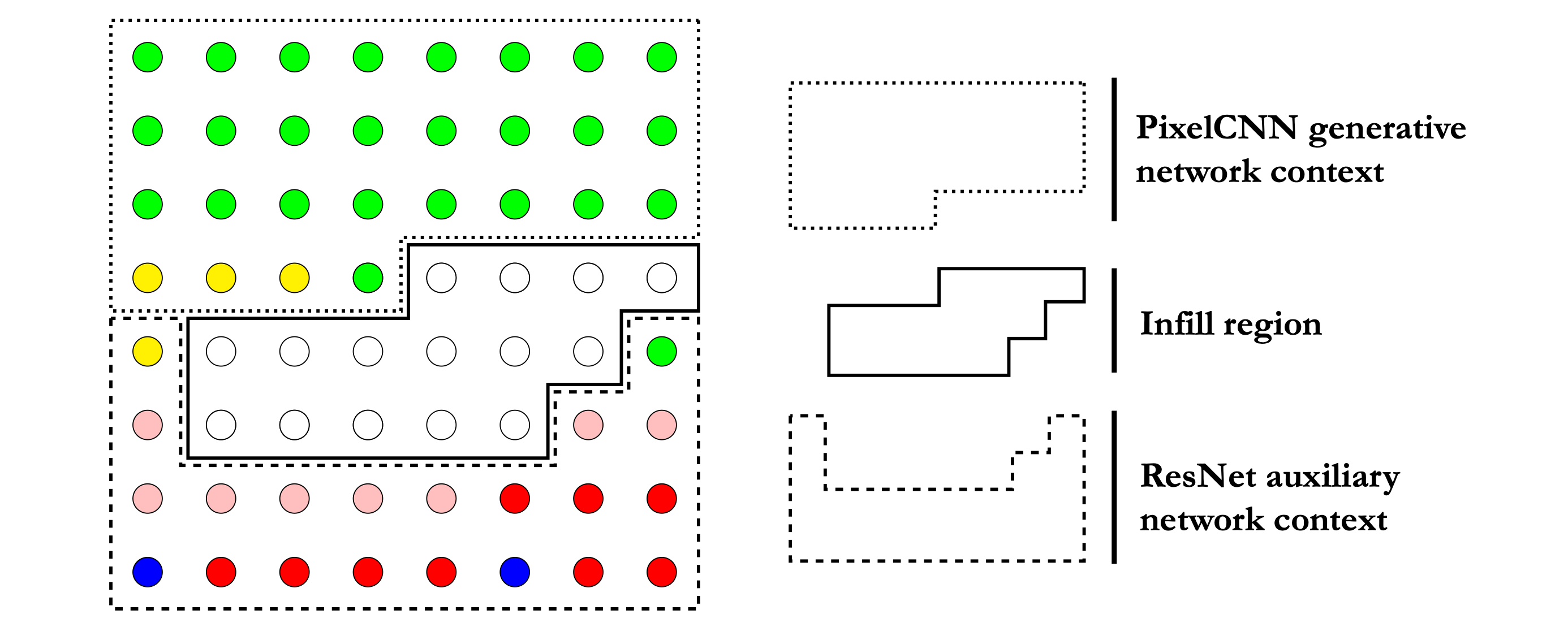}
    \caption{\textbf{Input regions for components of the \pccnn architecture}. The image completion spanning the white cells (middle region) above is sampled using a distribution modeled using an autoregressive PixelCNN network and an auxiliary ResNet network. The autoregressive network is only provided the pixels from the top region due to the design of its masked convolutional layers, while the ResNet auxiliary network uses pixels from the entire observed image (top and bottom regions). The prior context and conditional context for the single white pixel in row 4, column 5, counting from the left and top, are specified as shown.}
    \label{fig:regions}
\end{figure}

\begin{figure}[H]
    \centering
    \includegraphics[width=0.9\linewidth]{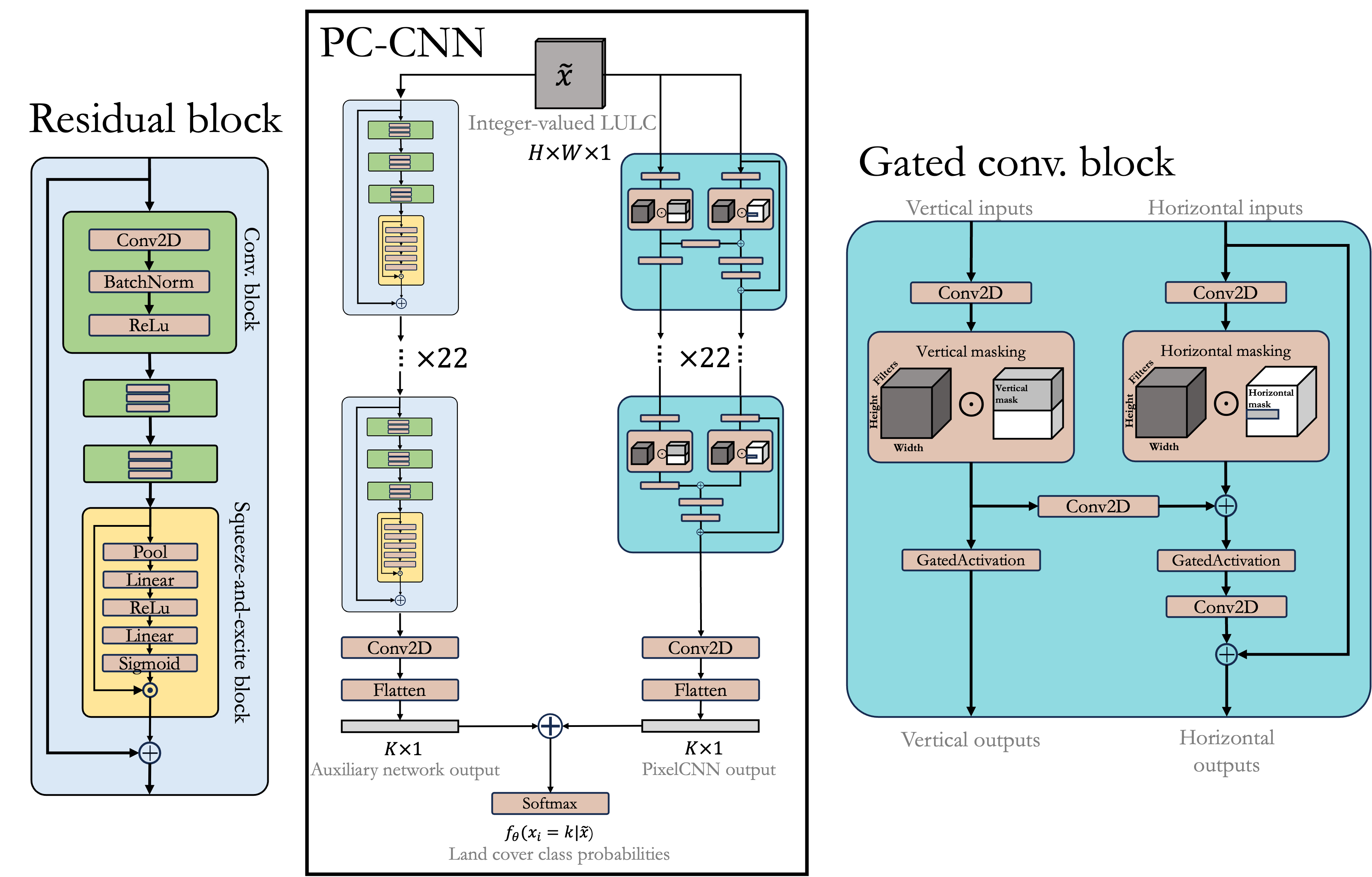}
    \caption{\textbf{Architecture for the \pccnn model.} The autoregressive PixelCNN model consists of gated convolutional blocks that use masks to prevent information leakage from future pixels during training. The auxiliary ResNet network conditions the PixelCNN model on the observed pixels and includes residual and squeeze-and-excite layers. Elementwise addition is indicated by $\oplus$ and multiplication by $\odot$.}
    \label{fig:architecture}
\end{figure}

In preliminary analyses for this work, we tried to develop an alternative modeling framework using a convolutional variational autoencoder \cite{kingma2014} along with Hamiltonian Monte Carlo \cite{duane1987} in the latent space of the autoencoder. Although this approach yielded acceptable results, it required about $10^5$ gradient evaluations per effective sample compared to $\approx 10^3$ forward model evaluations per sample with the \pccnn architecture. We used this architecture's implementation from \cite{dupont2019} with minor modifications. The architecture consists of a primary autoregressive generative network $f_{gen}$ and an auxiliary ResNet network $f_{aux}$, all implemented using residual connections between layers. These models are combined by adding their logits together via the equation $f_\theta(x_i | \tilde{\bm{x}}) = f_{gen}(\tilde{\bm{x}}) +  f_{aux}(\tilde{\bm{x}})$ where $\tilde{\bm{x}}$ denotes the partial observation of image $\bm{x}$. Figure \ref{fig:architecture} illustrates the types and number of layers and connections involved in both generative and auxiliary networks. We note that these terms correspond to the notions of prior and conditional networks in \cite{dupont2018}; we change this terminology to avoid confusion with the sampling performed in our analyses.

The core of the \pccnn architecture is an autoregressive PixelCNN network that takes an integer-valued input array $\tilde{\bm{x}}$ specified in two spatial dimensions and outputs a $K$-vector of per-class logits. This network design is based on gated convolutional blocks (Figure \ref{fig:architecture}) which use both vertical and horizontal masks to prevent blind spots in the model's receptive field. For more information on this phenomenon, we direct the reader to the original PixelCNN and PixelCNN++ studies \cite{vandenoord2016, salimans2017}. The primary operation in the gated convolutional block is a two-dimensional convolution, followed by elementwise multiplication with a binary mask array to zero out elements that would otherwise be visible to the network and correspond to future pixels in the raster scan order. The PixelCNN generative network comprises 22 of these blocks connected sequentially, followed by a final convolutional layer and flattening to a $K$-dimensional vector.

All convolutional layers have 96 filters and a kernel size of 5 pixels. We also use batch normalization and ReLU activation for both prior and auxiliary networks. The model totals 18.7 million parameters with 13.2M parameters in the autoregressive network and 5.5M parameters in the auxiliary network. For training, we utilize Adam with a learning rate of $5\times 10^{-4}$ and a batch size of 64 for 300 epochs, which takes about 225 hours on a Lambda Labs virtual machine equipped with an Nvidia A10 GPU, reaching log-likelihood values of 6.1 bits per dimension. All code and data required to reproduce our analyses are available at \texttt{github.com/ckrapu/nlcd-inpaint}.

\subsection*{Categorical conditional autoregression}
To understand the relative value of a deep approach compared to simpler, computationally cheaper alternatives, we implemented a benchmark spatial categorical conditional autoregression (\sccar) model with a categorical likelihood for comparison. We implemented this model to highlight the limitations of statistical models with fewer parameters that cannot capture complex correlation patterns that require extra parameters. This statistical model is an incremental extension of existing models such as spatial autoregressive multinomial models \cite{gelfand2003} as well as binary autoregressions \cite{geman1984}, and is closely related to ongoing research in models with spatial and cross-variable or class dependencies, such as \cite{dey2022}. In simple terms, the only innovation in this model is the integration of a cross-class correlation matrix $\bm{A}$ with per-class spatial correlation matrices $\bm{V}_k$. At a high level, this model handles the correlation structure between all $20\times1600$ per-pixel, per-class logits using a separable representation with distinct covariance matrices for cross-pixel and cross-class correlations. We selected this model as the simplest statistical model that permits spatial and cross-class correlations for data with unordered outcomes and finite support. This model represents the logits of the per-pixel LULC classes as draws from $K$ independent vectors with a conditional autoregressive prior, which are then subject to a linear transformation using $\bm{A}$. The rows of the matrix arising from this operation are vectors with both spatial correlations between the probabilities of the same class across neighboring sites and between different classes within the same site. Let $N$ denote the number of pixels per image; the model's prior specification is:

\begin{align*}
X_i & \sim \text{Categorical}(\bm{p}_i)  \\
p_{ik} &= \frac{e^{u_{ik}}}{\sum_{k'=1}^K e^{u_{ik'}}} \\
\bm{U} &= \bm{\Omega} \cdot \bm{A} \\
\bm{\omega}_{\cdot k} &\sim \text{Normal}\left(m_k \bm{1}_N, \bm{V}\right) \\
\bm{V}_k &= \tau_k \cdot D (\bm{I}_N - \rho_k D^{-1}Q) \\
\rho_k &\sim \text{Uniform}(0,1) \\
m_k &\sim \text{Normal}(0, 2) \\
\tau_k &\sim \text{Cauchy}^+(\beta=2) \\
\bm{A} &\sim \text{LKJ}(\eta = 1)
\end{align*}

Where $X_i$ is the land cover class at pixel $i$, $\bm{P}$ is an $N \times K$ matrix of per-pixel, per-class probabilities, and $\bm{U}$ is an identically-sized matrix of real-valued logits. The $\text{Cauchy}^+$ refers to the truncated Cauchy distribution with mode at zero and $p(\sigma) = 0$ for all negative $\sigma$. $\bm{1}_N$ is an $N$-dimensional vector of ones, $\bm{I}_N$ is an $N\times N$ identity matrix, $\bm{Q}$ is an $N \times N$ adjacency matrix shared among all CAR vectors, while $\bm{D}$ is an $N\times N$ diagonal matrix with elements set to the row sums of $\bm{Q}$, indicating the number of neighbors per element. The Lewandowski-Kurowicka-Joe (LKJ) prior over correlation matrices \cite{lewandowski2009} takes a single parameter $\eta$; when $\eta=1$, the prior is uniformly distributed over valid correlation matrices. Conceptually, the matrix $\bm{A}$ provides correlations between different spatial vectors, and the vectors $\bm{\omega}_{\cdot, k}$ are spatially correlated through the dependence of their covariance matrix on the adjacency matrix $\bm{Q}$. The autocorrelation parameters $\rho_k$ determine the strength of this spatial correlation for each vector. By allowing for multiple autocorrelation parameters, we can model spatial patterns of land cover occurrence with varying degrees of spatial compactness. For instance, pixels of open water like lakes and large rivers tend to show extremely high spatial correlation, while regions of low-intensity urban development interspersed with cropland and forest exhibit lower autocorrelation values.

\begin{figure}[H]
\centering
\begin{tikzpicture}

  \node[obs] (x) {$\bm{x}_{ik}$};
  \node[latent, right=of x] (p) {$p_{ik}$};
  \node[latent, above=of p] (U) {$U_{ik}$};
  \node[latent, left=of U] (omega) {$\omega_{ik}$};
  \node[latent, right=of U] (A) {$A_{kk'}$};
  \node[latent, above=of omega] (V) {$V_k$};
  \node[latent,  left=of V] (rho) {$\rho_k$};
  \node[latent, right=of V] (m) {$m_k$};
  \node[latent, above=of m] (sigma) {$\sigma$};
  \node[latent, below left=of V] (tau) {$\tau_k$};
  \node[obs, above=of V] (D) {$\mathbf{D}$};

  \draw[->] (p) -- (x);
  \draw[->] (U) -- (p);
  \draw[->] (omega) -- (U);
  \draw[->] (A) -- (U);
  \draw[->] (V) -- (omega);
  \draw[->] (rho) -- (V);
  \draw[->] (sigma) -- (m);
  \draw[->] (tau) -- (V);
  \draw[->] (m) -- (omega);
  \draw[->] (D) -- (V);

  \plate  {plateL} {(x) (p) (U) (omega)} {$N$}
  \plate [inner xsep=8pt, inner ysep=8pt] {plateK} {(x) (p) (U) (omega) (V) (rho) (m) (tau) (A)} {$K$}
  \plate  {plateKprime} {(A)} {$K'$}

\end{tikzpicture}
\caption{\textbf{Plate diagram of the benchmark latent conditional autoregressive model}. The observed adjacency matrix $\bm{D}$ informs the spatial covariance of $\omega_{\cdot k}$. Cross-class correlations between elements of $\bm{U}$ are induced via the matrix $\bm{A}$. Shaded circles represent observed quantities while unshaded circles denote parameters to estimate. $N$ is the number of pixels in the image, and $K$ is the number of land cover classes. $K'$ refers to the second index running over ${1,...,K}$, which is only used to index the columns of the matrix $\bm{A}$.}
\label{fig:graphical}
\end{figure}
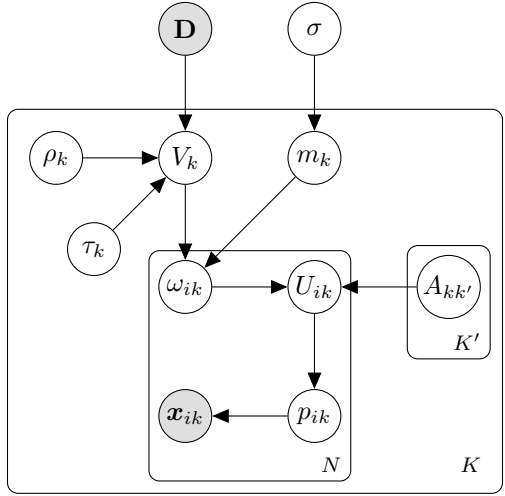

Each entry $q_{ij}$ in $\bm{Q}$ is defined as
  \[ q_{ij} = 
  \begin{cases} 
  1 & \text{if } i \text{ and } j \text{ are 4-adjacent pixels} \\
  0 & \text{otherwise} 
  \end{cases} 
  \]
We utilize an implementation of the CAR distribution in PyMC \cite{salvatier2016} which allows an $\mathcal{O}(N)$ evaluation of the density of $\omega$ through sparse operations involving the matrix $\bm{D}$. Defining $\bm{T} = \text{diag}(\tau_1,...,\tau_K)$ and $C=D (I - \rho D^{-1}Q)$, we can express the prior for $\bm{U}$ as a matrix-normal distribution with covariance matrices $TC$ and $AA^T$. Ignoring constant terms for the densities of $\rho_k$ and $\bm{A}$, the log posterior density of the parameters under this model is given by:
\begin{align*}
    \log p(\bm{\Omega}, \bm{A}, \bm{m}, \bm{\tau}, \bm{\rho} | \bm{x}) &\propto \sum_i \sum_{k=1}^{K} I(X_i = k) \log \left( \frac{e^{\sum_j \omega_{ij} A_{jk}}}{\sum_{k'=1}^K e^{\sum_j \omega_{ij} A_{jk'}}} \right) \\
    &\quad - \sum_k \left[ \frac{1}{2} \log |\tau_k D (\bm{I}_N - \rho_k D^{-1}Q)| + \frac{1}{2} (\bm{\omega}_{\cdot k} - m_k \bm{1}_N)^\top [\tau_k D (\bm{I}_N - \rho_k D^{-1}Q)]^{-1} (\bm{\omega}_{\cdot k} - m_k \bm{1}_N) \right] \\
    &\quad - \sum_k \frac{1}{4} m_k^2 + \sum_k \log\left(\frac{2}{\pi(4 + \tau_k^2)}\right)
    \end{align*}    
The \sccar model is not suitable for fitting the entire training data set directly due to the number of free parameters scaling as $\mathcal{O}(K(K-1)/2 + 4K)$ and $\mathcal{O}\left(N(K-1)\right)$ latent variables per image; instead, we use it on a single, partially-occluded image for comparison in section \ref{section:case_study}. A graphical model diagram is shown in Figure \ref{fig:graphical}. To fit this model, we employed the built-in No-U-Turn sampler \cite{hoffman2013}, running four chains with 2000 tuning samples and 2000 samples for evaluation with a target acceptance rate of 0.9. All $\hat{R}$ values \cite{gelman1992} were below 1.05. Samples were generated via ancestral sampling, first drawing from the posterior distribution over parameters and then from the implied distribution of LULC probabilities for missing pixels. Several such draws are presented in Figure \ref{fig:comparison}.

\begin{figure}[H]
    \centering
    \includegraphics[width=0.7\linewidth]{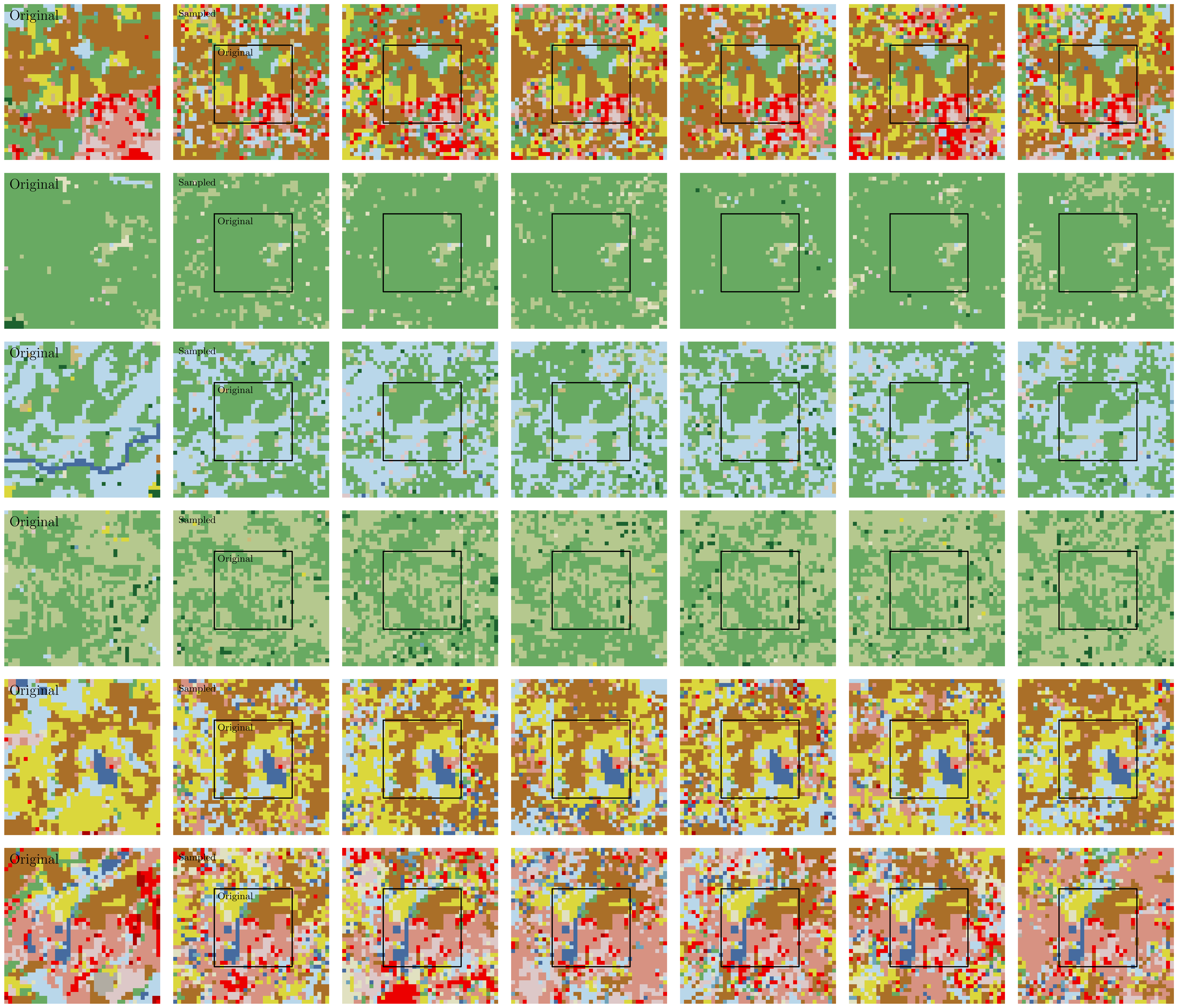}
    \caption{Image completions using the spatial categorical conditional autoregressive model. These completions help to highlight spatial correlation patterns over a larger area.}
    \label{fig:sccar}
\end{figure}

We note several key differences between the \sccar and \pccnn models that make a direct comparison unreasonable. The former has far fewer parameters and represents the per-class logits as linear functions of underlying latent variables $\omega$ without directly modeling spatial correlations beyond neighboring pixels. Its relatively simple structure, therefore, benefits less from a large training dataset. We also considered using multipoint statistical methods suitable for categorical data \cite{deutsch2006}, but found no satisfactory solution to this approach's need for a distance function between image patches that could accommodate larger values of $K$ without becoming impractical. In this situation, identifying suitable image patches for infill required evaluating an unacceptable number of candidates.

\section{Analyses \& Results}
We conduct several analyses focused on different aspects of using a deep autoregressive model compared to a simpler statistical model. Each of these is described in more detail below. 
\subsection{Unconditional sampling}
\begin{figure}[H]
    \centering
    \includegraphics[width=0.6\linewidth]{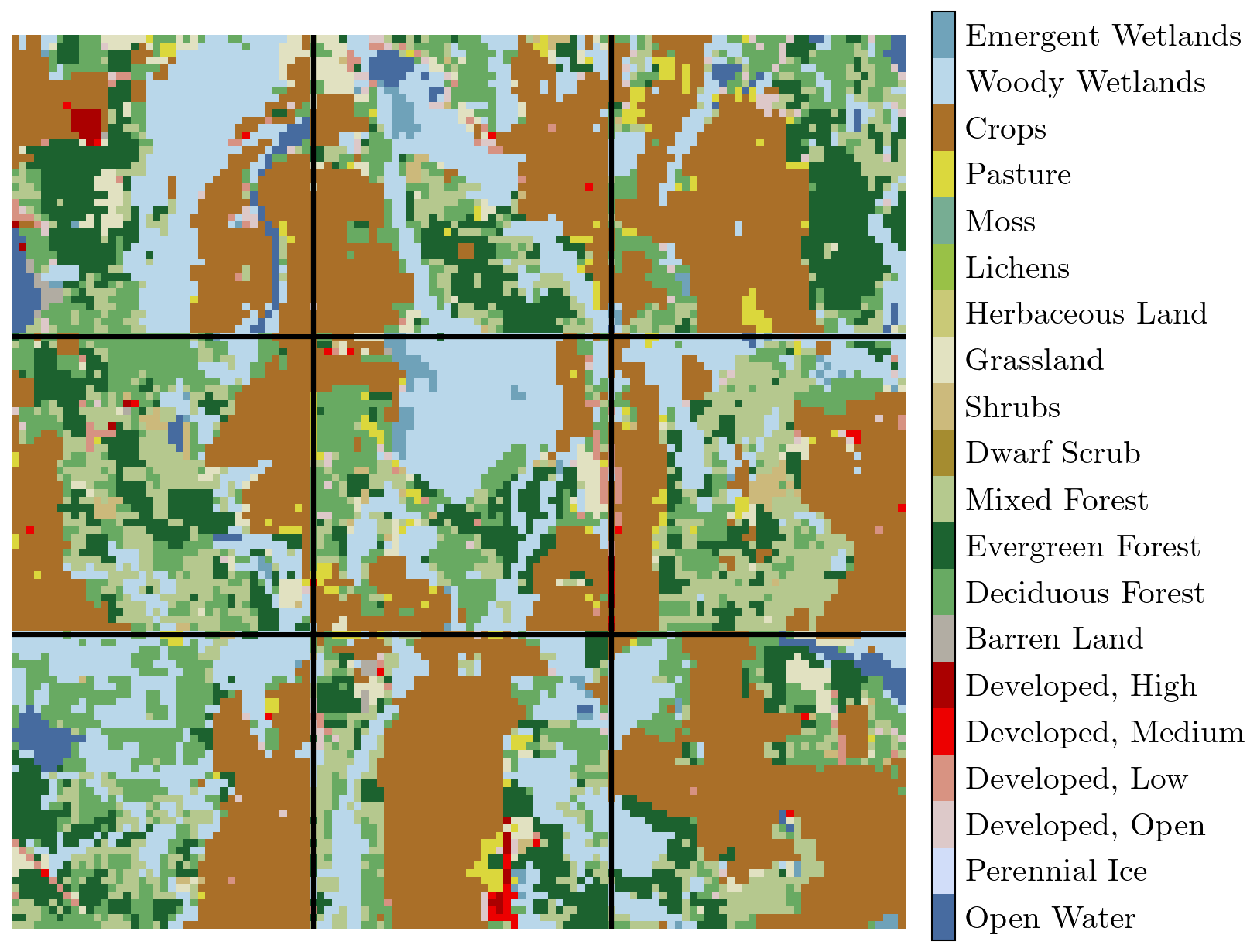}
    \caption{\textbf{Unconditional samples drawn from \pccnn model}. These images were generated by passing a fully-masked image to the generative model.}
    \label{fig:unconditional}

\end{figure}

While the main focus of this work is conditional sampling, i.e., infilling partial images given a subset of visible pixels, we are also interested in understanding whether the \pccnn model can generate images through unconditional sampling. A random selection of nine samples is shown in \ref{fig:unconditional}. Although there is substantial variation in orientation, size, and types of LULC patterns represented, the overall proportion of pixels in each class seems underdispersed compared to the training data as most unconditional samples are a mixture of cropland, forest, and wetland. While these samples resemble the training data, they do not adequately represent simpler images with most or all pixels in a single LULC type; this latter class of images is common in the training data.

\subsection{Image inpainting}
\begin{figure}[H]
    \centering
    \includegraphics[width=1.2\linewidth]{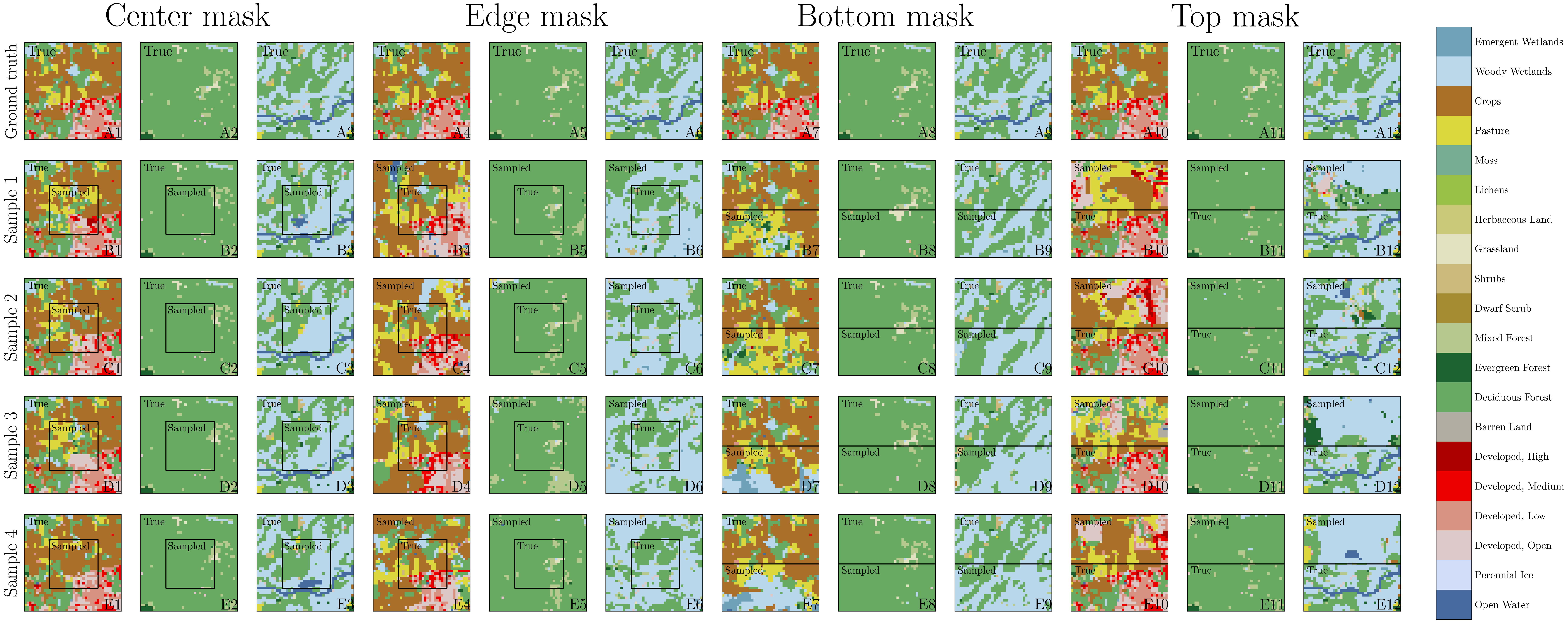}
    \caption{Sampled image completions across three different masking arrangements}
    \label{fig:completions}
\end{figure}

Figure \ref{fig:completions} shows four image completions on held-out data for three different images and three different masking scenarios. We notice major discrepancies between completions produced under the bottom and top masking schemes; in the former case, the top of the image is used to drive the autoregressive primary network while the auxiliary network contributes relatively little since it focuses on the out-of-order pixels which are already supplied to the autoregressive network. When completing the bottom half of the image, linear features such as roads are easily generated; the sixth column in Figure \ref{fig:completions} illustrates this clearly. However, when the top part of the image is occluded, the generative process relies much more on the auxiliary network, which struggles to complete road features, as shown in the rightmost column of Figure \ref{fig:completions}. Nevertheless, in all three masking schemes, a diverse range of image completions is generated, regardless of whether the prior or auxiliary network is more directly involved in the generation. Furthermore, the model does not have an intrinsic dependence on image orientation; we can rotate the conditional context to allow the autoregressive network access to observed pixels.

An issue raised in \cite{dupont2019} was the relative difficulty in making the generative model honor the conditioning information for out-of-order pixels for some model architectures. We observe this issue in the second column of generated completions in \ref{fig:completions}. Here, despite the auxiliary network having access to observed pixels, these true pixels are overlooked in the completed image, resulting in an image that does not match the ground truth, even on the pixels that are not masked.

\subsection{Diversity \& calibration}
\begin{figure}[H]
    \centering
    \includegraphics[width=0.8\linewidth]{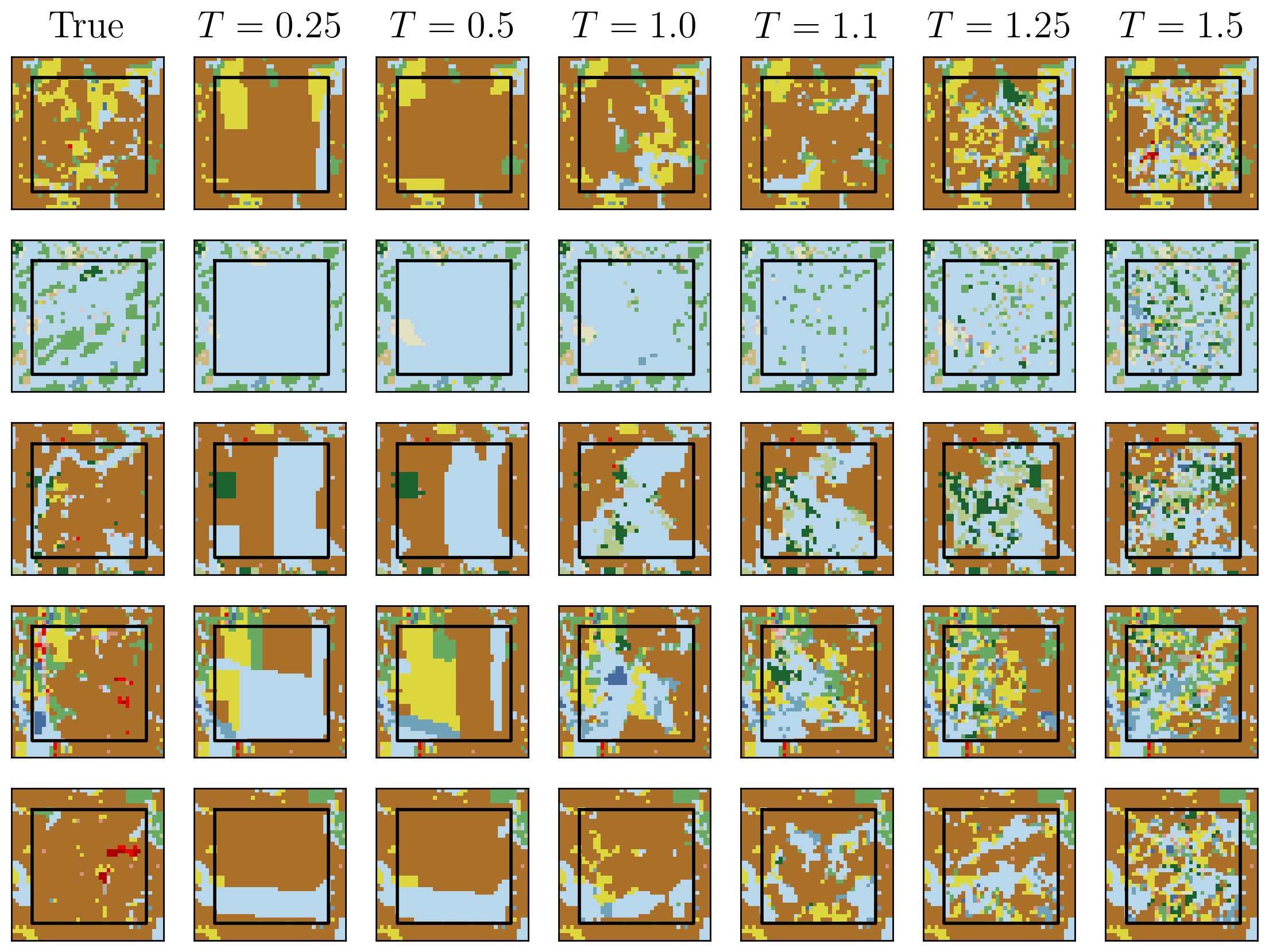}
    \caption{\textbf{Image completion using different temperature values}. Values inside the black box were produced by the \pccnn model. Low temperature values tend to produce overly homogeneous LULC completions, while high values lead to weaker spatial correlation patterns.}
    \label{fig:temperature}
\end{figure}

The primary intent of this work is to assess autoregressive image completion for LULC, and we would like to know whether the predictive distribution over spatial patterns is \emph{reliable} in the sense that coverage probabilities for credible intervals of selected summary statistics of infilled land use maps are close to their nominal values. This criterion circumvents issues with producing objective measures of image diversity using secondary machine learning models \cite{zhang2018, zheng2019}. Our summary statistics are image entropy, the spatial join statistic, the number of 4-connected contiguous image patches, and the proportion of pixels falling into the image's modal class. These functions are defined in Table \ref{tab:statistics}. We repeated these analyses with final-layer softmax temperature parameters in the set $\{0.25, 0.5, 1.0, 1.25, 1.5\}$ to understand whether this parameter can be used to tune the diversity and calibration of the distribution of sampled images. We did not adjust the temperature hyperparameter during training. Figure \ref{fig:temperature} illustrates the effect of varying the softmax temperature on image completion across several examples.

\begin{table}[H]
    \centering
    \begin{tabular}{|ll|}
    \hline
    \textbf{Description}& \textbf{Formula} \\
    \hline
     Image entropy  & $H(\bm{x}) = -\sum_{k=1}^K \frac{\sum_{i}I(x_i=k)}{N} \log\left(\frac{\sum_{i}I(x_i=k)}{N}\right)$ \\
     Adjacency statistic  & $a(\bm{x})=\sum_{i=1}^{N} \sum_{j \in \text{adj}(i)} I(x_i = x_j)$ \\
     Number of 4-Connected Patches  & $n_{connected}(\bm{x})=\left| \bigcup_{(i, j) \in \mathcal{P}} \text{Region}_{i,j}(\bm{x}) \right|$ \\
     Proportion in modal class & $p_{modal}(\bm{x}) = \frac{1}{N} \sum_{i=1}^{N} I\left(x_i = \textrm{mode}(\bm{x})\right)$\\
    \hline
    \end{tabular}
    \caption{\textbf{Summary statistics for image analysis}. The function Region$(\cdot)$ denotes the set of pixels in the same contiguous region sharing the same class label.}
    \label{tab:statistics}
    \end{table}
    
\begin{table}[H]
    \centering
    \setlength{\tabcolsep}{4pt} 
    \begin{tabular}{p{2.5cm}lcccccc}
    \toprule
     &  & \multicolumn{6}{c}{Coverage Rate} \\
    \parbox{2.5cm}{\centering Statistic} & \parbox{1.5cm}{\centering Percentile} & $T=0.25$ & $T=0.5$ & $T=1.0$ & $T=1.1$ & $T=1.25$ & $T=1.5$ \\
    \midrule
    \multirow[t]{3}{=}{Image entropy} & 50 & 0.05 & 0.10 & 0.10 & 0.10 & 0.20 & 0.00 \\
     & 90 & 0.20 & 0.25 & 0.30 & 0.40 & 0.35 & 0.00 \\
     & 95 & 0.25 & 0.30 & 0.45 & 0.40 & 0.35 & 0.00 \\
    \cline{1-8}
    \multirow[t]{3}{=}{Adjacency statistic} & 50 & 0.00 & 0.00 & 0.20 & 0.20 & 0.10 & 0.00 \\
     & 90 & 0.00 & 0.00 & 0.40 & 0.50 & 0.30 & 0.00 \\
     & 95 & 0.00 & 0.00 & 0.45 & 0.60 & 0.30 & 0.00 \\
    \cline{1-8}
    \multirow[t]{3}{=}{Proportion in modal class} & 50 & 0.10 & 0.15 & 0.20 & 0.20 & 0.00 & 0.10 \\
     & 90 & 0.15 & 0.30 & 0.50 & 0.40 & 0.30 & 0.10 \\
     & 95 & 0.30 & 0.35 & 0.60 & 0.40 & 0.40 & 0.10 \\
    \cline{1-8}
    \multirow[t]{3}{=}{Patch count} & 50 & 0.00 & 0.00 & 0.00 & 0.15 & 0.10 & 0.00 \\
     & 90 & 0.00 & 0.00 & 0.05 & 0.30 & 0.35 & 0.10 \\
     & 95 & 0.00 & 0.00 & 0.05 & 0.30 & 0.45 & 0.15 \\
    \cline{1-8}
    \bottomrule
    \end{tabular}
    \caption{\textbf{Nominal and actual coverage rates for \pccnn image completions assessed across several statistics}. Due to the slow sampling speed, only 20 images were included with 100 samples per image, leading to coverage levels restricted to the values $0\%, 5\%, ..., 95\%, 100\%$.}
    \label{tab:coverage_rates}
    \end{table}
    
We find that for all statistics and all temperature values, the predictive distribution of image infill patches is significantly narrower than the actual distribution, when assessed using the summary statistics described earlier (Table \ref{tab:coverage_rates}). The patch count statistic distribution is particularly underdispersed with only one or no test images having a true value within the predictive interval. Increasing the temperature from $1.0$ to $1.1$ helps considerably but does not achieve full alignment between nominal and actual coverage probabilities. We also observe that the distribution at $T=1.0$ appears to be better than that at $T=1.1$ for the modal class proportion at the $90\%$ and $95\%$ probability levels, although this is not the case for other statistics and probabilities. While there is extensive literature on calibration for scalar-valued classification tasks \cite{guo2017}, little guidance is available on how to achieve this for prediction problems that are inherently high-dimensional, involving generative models.

\subsection{Comparison with \sccar model}
\begin{figure}[H]
    \centering
    \includegraphics[width=0.9\linewidth]{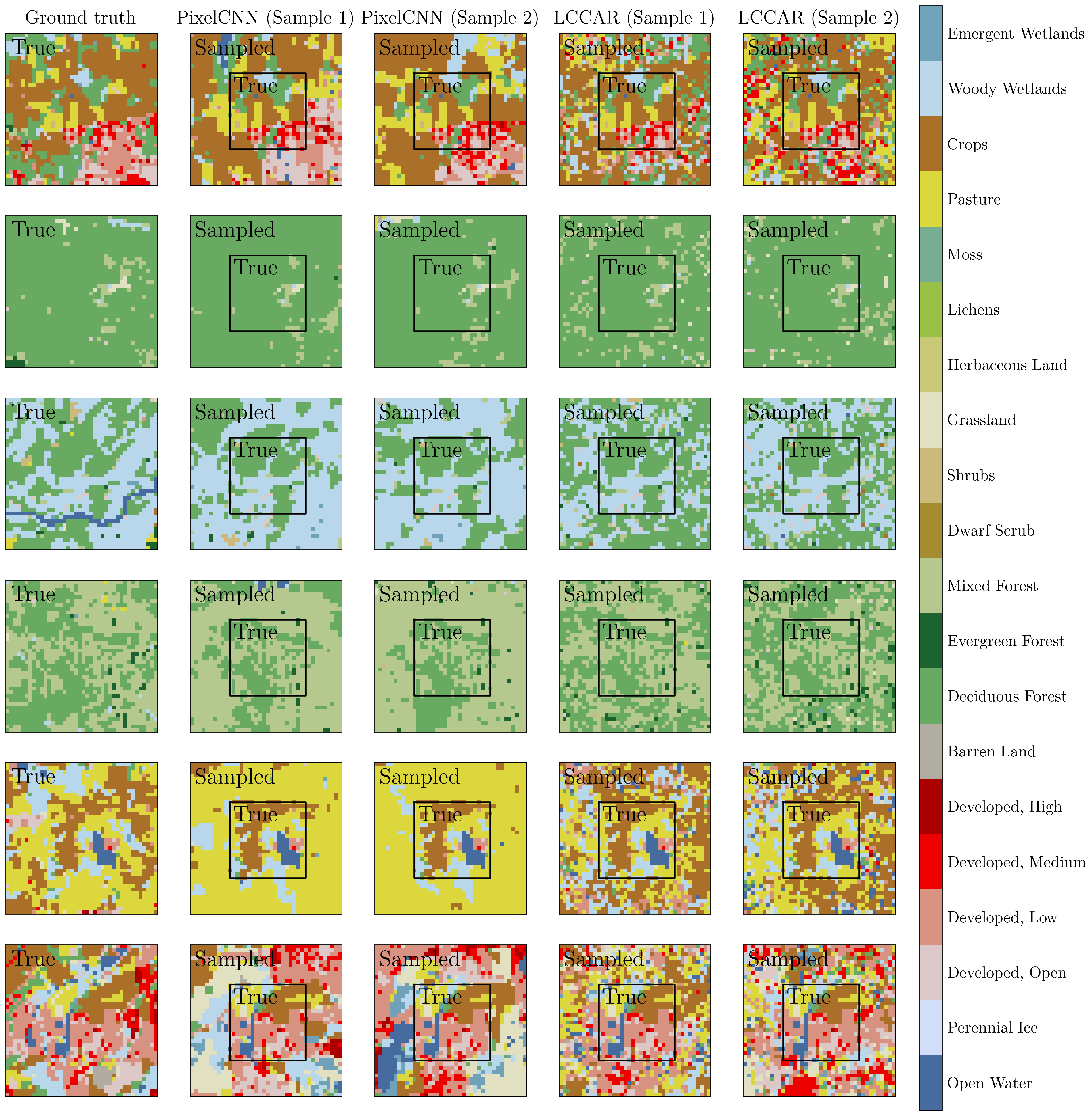}
    \caption{\textbf{Comparison of image completions between \pccnn and \sccar models.} Samples drawn from the \sccar model show less realistic LULC patterns with frequent interspersal of different land cover classes and lower levels of spatial contiguity than the samples produced from the \pccnn model.}
    \label{fig:comparison}

\end{figure}

To help illustrate the need for highly parameterized models of LULC data, we compare conditional samples from the \pccnn model with those from the \sccar model in Figure \ref{fig:comparison} for five different images, each with a dominant land cover type.  We find that samples from the \sccar model generally show relatively low levels of spatial cohesion and are inadequate for modeling large areas of cropland, forest, or other classes. This is likely due to the relatively small number of parameters in the \sccar model which cannot represent the complex spatial patterns present in the training data. In contrast,  the \pccnn model is able to produce samples visually similar to the training data that show high levels of spatial contiguity.

\section{Case study} \label{section:case_study}
For a variety of use cases, employing an autoregressive model for inference using only the imagery fitting within the model's immediate spatial context would not appropriate. For example, we may wish to produce an infill for a region much larger than the size of the images used for training. As a first step toward solving this problem, we attempt to use a solution whereby we sequentially fill in small portions of a larger whole, using the previously simulated pixels as conditioning for subsequent applications of the \pccnn model. 

We demonstrate this approach on an area centered on the coordinates 42.321 N., 85.318 W., focused on a Michigan state park formed from land previously administered by a nearby military base. Our case study is framed around the task of determining the likelihood of development on this land had the military base and state park not been established.

\begin{figure}[H]
    \centering
    \includegraphics[height=0.7\textheight]{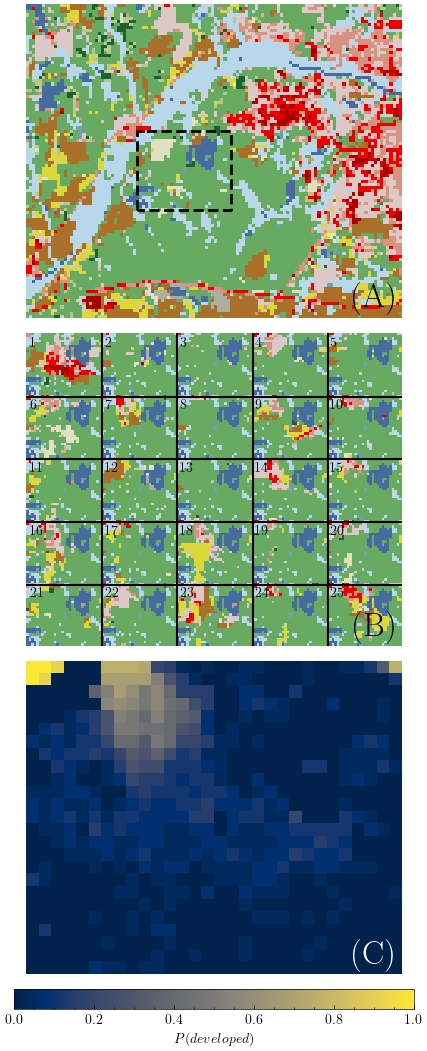}
    \caption{\textbf{Samples of land cover completion for case study site}. The area outlined in (A) was masked and then filled in using the \pccnn model. As the masked region was significantly larger than the training data size, the infill was conducted sequentially using surrounding pixels to provide spatial context. 25 infill completions are shown in (B), with (C) displaying the proportion of samples across which a pixel was assigned to one of the four developed land use classes. The influence of nearby urban regions proximal to the northeast and northwest corners of the infill region induce high probabilities of development in the upper left and right parts of the domain (C).}
    \label{fig:case-study}
\end{figure}

The procedure used for infilling a larger region is as follows: we find the top-most pixel in the image which is missing, and then extract a window of surrounding context of up to 27 pixels above and to the left so that roughly 1/9 of the window is missing for each infilling. We then use the model to predict the missing pixels, using these infills as conditioning for the next infill. We repeat this process until the entire image is filled in. At each iteration, we randomly flip the image vertically and horizontally to avoid biasing the model toward any particular orientation. Depending on the order of infill, each full simulation required either 5 or 6 separate infills to be produced (Figure \ref{fig:case-study}).

A major limitation of this approach is that any spatial information at ranges of 27 pixels or farther would be discarded. Attempts to address this would likely involve adjustments to model architecture and training procedures to allow for long-range information to be included in a computationally efficient manner. A consequence of the former issue is that for missing regions with dimensions larger than the training data size, the model may generate partial infills which are consistent with the immediate spatial context but which jointly are not reasonable. For example, if a missing region covers a single large parcel of land owned entirely by one entity, an ideal infill would be one class label for the entire region. However, the model may generate a patchwork of different class labels if the region is much larger than 40 pixels in either spatial dimension.
As shown in Figure \ref{fig:case-study}, the modeled image completions include varying degrees of urban development, pasture, cropland, and forest. We find that the per-pixel proportion of samples showing development at that location smoothly varies in space, ranging from $p=1.0$ in the northwest corner of the study area to nearly 0.0 in the southern sector of the study area. This provides a plausible explanation that the area showing high likelihood of development is adjacent to existing development, and areas which are separated from existing development by water features (northeast corner) are likely to remain undeveloped.

\subsection*{Geographic variation in modeled likelihood}
\begin{figure}[H]
    \centering
    \includegraphics[width=\linewidth]{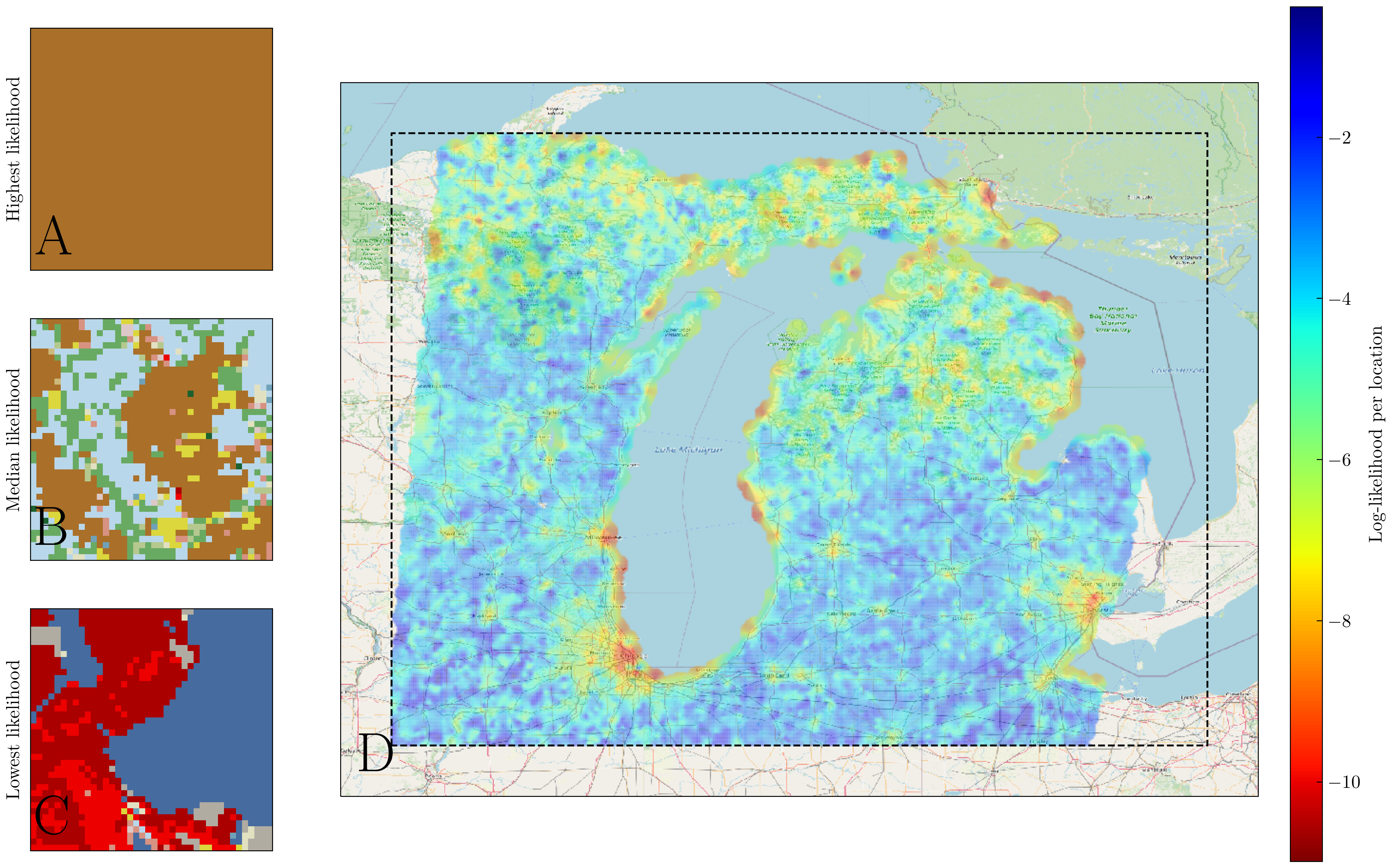}
    \caption{\textbf{Map of per-image likelihood under the \pccnn model} Coloring on the map results from a radial basis function interpolation of per-image likelihood values. Warmer colors correspond to locations with LULC images assigned lower model likelihoods. Images with a single land cover class (A) where the land cover class is common yield high likelihood values, while uncommon combinations, such as high urban development next to open water, yield lower values.}
    \label{fig:likelihood}

\end{figure}

To understand the \pccnn model's tendency to generate certain types of images or land cover patterns, we scored each image in the test dataset using the model to produce a per-location log-likelihood value which was then interpolated using radial basis functions as shown in Figure \ref{fig:likelihood}. A pattern that emerges is that urban- and water-adjacent areas tend to receive lower likelihood values; this is likely due to their scarcity in the training data compared to cropland and forest. As we discarded any training images with more than 50\% of its pixels in the \emph{open water} class, this LULC type is infrequent in the training data despite the geographic area covering several of the Great Lakes. Images from the vicinity of Chicago, Illinois and Detroit, Michigan received lower likelihood values than surrounding areas. The minimum likelihood value attained by any image in the test set corresponds to an image with high urban density next to open water. Large areas of high likelihood values were noted in the rural interior of Wisconsin, Illinois, and Michigan. Areas with extensive forest and mixed land use in the northern part of the study area received likelihood values in the middle of the range.

\section{Discussion}

\subsection*{Advantages}
While acknowledging significant limitations in empirical coverage rates, we identify several advantages to using a deep autoregressive network for LULC modeling. The \pccnn model exhibits the ability to capture complex spatial correlation patterns characteristic of LULC data, which traditional spatial statistical models often find difficult. The model's flexibility, due to its millions of parameters, allows for the generation of outputs that are not only diverse but also rich in detail. It can be calibrated against large datasets, leveraging the abundance of available geospatial data to improve its predictive accuracy. Furthermore, there is a clear path to incorporating additional data sources by extending this single-channel autoregressive model to include multiple channels, as seen in the computer vision literature for natural images \cite{vandenoord2016}.
\subsection*{Limitations}
By solely relying on spatial land cover data without ancillary datasets such as elevation, climate variables, or economic information, this approach cannot exploit strong patterns of cross-correlation across data layers, i.e. forested hillsides or high development density near population centers. Current leading approaches to performing LULC forecasting and change \cite{sohl2014} often make use of multiple data sources for increased effectiveness. In particular, representing LULC in discrete parcels is crucial for producing realistic LULC patterns and is extensively used in ongoing LULC forecasting efforts \cite{sohl2019}. For including additional process detail, agent-based modeling frameworks \cite{kaiser2020} may prove useful. Concerning limitations of the specific network architecture chosen, we recognize that this approach shares the difficulties in capturing long-range correlations as noted in \cite{chen2018}. Moreover, sampling is relatively slow compared to single-pass methods such as variational autoencoders \cite{razavi2019} and generative adversarial networks \cite{brock2018}.

We observe that the predictive intervals of per-image summary statistics produced by the distribution of completed images are generally too narrow, with the actual coverage rates falling well below the intended rates. In future work, we plan to explore whether automatic tuning of the model's temperature parameter can help remedy this lack of image diversity and calibration.

\subsection*{Future work}

\subsubsection*{Alternative model architectures}
A significant number of improvements in the area of autoregressive modeling have emerged since the development of the original PixelCNN model. One of the shortcomings of this model is its inefficiency at representing long-range correlations due to the required masking arrangement. The PixelSNAIL model developed in \cite{chen2018} addresses this issue with the inclusion of a self-attention module that lacks a finite receptive field like most convolutional layers and instead allows for activations to depend on nonlinear functions of pixels on opposite sides of an image. While the PixelCNN++ model \cite{salimans2017} is indicated as a successor to PixelCNN, several of the modifications pursued in that work are of little relevance for LULC modeling. The mixture-of-logistics approach used in that work to encourage the K-outcome categorical distribution to behave more like a discretization of a continuous distribution over an ordered set of numbers is inappropriate for our application as our per-pixel classes lack any natural ordering. Furthermore, that work's exploration of conditioning on subpixel information is irrelevant here as we use only a single input channel instead of RGB data. With recent interest in models leveraging generative pretraining with autoregressive data for text, investigations into using the same architectures for 2D image data \cite{chen2020} have also found encouraging results, presenting a closely-related line of models appropriate for further investigation. For LULC modeling, a larger image size is highly desirable as it allows for more spatial context to be used for conditioning. However, recent advances in the synthesis of high-quality, large images using latent diffusion models \cite{rombach2022} may be difficult to apply directly to this problem as the underlying diffusion process assumes real-valued image data, although there are efforts to extend the approach to discrete data \cite{chen2023}. Finally, we have not experimented with or prototyped any image completion strategies using flow-based methods for discrete data \cite{tran2019} which may offer a viable alternative. We are also interested in exploring alternative model structures requiring different data representations. For example, representing parcels of land by polygons with associated land cover classes may be far more computationally efficient when used with a token-based autoregressive architecture similar to \cite{birsak2022}.

\subsubsection*{Additional data sources}
While large datasets of natural images commonly used in computer vision lack auxiliary data, data arising in a geospatial context often include the collocation of multiple data layers such as elevation \cite{jarvis2008}, climate \cite{fick2017}, soil type, and points of interest. Future research could benefit from the inclusion of these auxiliary layers. Specifically, investigating how LULC assignment between cropland and pasture correlates with soil type or how the likelihood of urbanization changes with proximity to roads or infrastructural features could be of interest. 

\subsubsection*{Extending contextual window}
The current approach to infilling larger regions is limited by the size of the training data. In particular, we are unable to utilize information outside of the $40\times40$ pixel window used for training. We identify several approaches to help address this issue. A straightforward way to achieve this is to include an additional modeling component that can take a lower-resolution summary of either surrounding land cover proportions or coarser imagery in a larger spatial window. A second approach is to use geographic vector embeddings \cite{wozniak2021} to capture information about nearby regions in a relatively compact manner. Lastly, we could employ a hierarchical model \cite{vahdat2021, defauw2019} to capture information at multiple scales.

\section{Conclusion}
Deep learning and autoregressive models have enabled the generation of realistic-looking patterns of land use and land cover given sufficient training data. Furthermore, we find that the discrete-output autoregressive models like the \pccnn model are suited to the task of generating and inpainting categorical data. However, we identify two challenges to such an approach in real-world workflows: (1) an under-diverse population of sampled image completions, (2) relatively slow sampling speed, limiting the application of this approach to large-scale image imputation problems. In future work, we plan to explore methods for addressing both of these challenges through modification of network design and further tuning over hyperparameters. With these issues resolved, we expect this modeling approach to be a practical tool for quantifying the impact of LULC change by providing illustrative reference distributions of counterfactual land use patterns.

\section*{Acknowledgments}
This work was supported by the U.S. Department of Defense Strategic Environmental Research and Development Program (project RC20-3054).

\printbibliography
\end{document}